\def\year{2020}\relax
\documentclass[letterpaper]{article} % DO NOT CHANGE THIS
\usepackage{aaai20}  % DO NOT CHANGE THIS
\usepackage{times}  % DO NOT CHANGE THIS
\usepackage{helvet} % DO NOT CHANGE THIS
\usepackage{courier}  % DO NOT CHANGE THIS
\usepackage[hyphens]{url}  % DO NOT CHANGE THIS
\usepackage{graphicx} % DO NOT CHANGE THIS
\usepackage{graphicx}  % DO NOT CHANGE THIS
\usepackage{amsmath}
\usepackage{amssymb}
\usepackage{booktabs}
\usepackage{lipsum}

\title{ZoomNet: Part-Aware Adaptive Zooming Neural Network for 3D Object Detection}
\author{
Zhenbo Xu\textsuperscript{\rm 1},
Wei Zhang\textsuperscript{\rm 2},
Xiaoqing Ye\textsuperscript{\rm 2},
Xiao Tan\textsuperscript{\rm 2},
Wei Yang\textsuperscript{\rm *1},
Shilei Wen\textsuperscript{\rm 2},
Errui Ding\textsuperscript{\rm 2},\\
\bf \Large Ajin Meng\textsuperscript{\rm 1},
Liusheng Huang\textsuperscript{\rm 1}
\\
\textsuperscript{\rm 1}University of Science and Technology of China\\
\textsuperscript{\rm 2}Department of Computer Vision Technology (VIS), Baidu Inc., China\\
\textsuperscript{*}Corresponding Author. E-mail: qubit@ustc.edu.cn
}

\begin{document}
\maketitle
\begin{abstract}
% 2000 words

3D object detection is an essential task in autonomous driving and robotics. Though great progress has been made, challenges remain in estimating 3D pose for distant and occluded objects. In this paper, we present a novel framework named ZoomNet for stereo imagery-based 3D detection. The pipeline of ZoomNet begins with an ordinary 2D object detection model which is used to obtain pairs of left-right bounding boxes. To further exploit the abundant texture cues in rgb images for more accurate disparity estimation, we introduce a conceptually straight-forward module -- adaptive zooming, which simultaneously resizes 2D instance bounding boxes to a unified resolution and adjusts the camera intrinsic parameters accordingly. In this way, we are able to estimate higher-quality disparity maps from the resized box images then construct dense point clouds for both nearby and distant objects. Moreover, we introduce to learn part locations as complementary features to improve the resistance against occlusion and put forward the 3D fitting score to better estimate the 3D detection quality. Extensive experiments on the popular KITTI 3D detection dataset indicate ZoomNet surpasses all previous state-of-the-art methods by large margins (improved by 9.4\% on AP\textsubscript{bv} (IoU=0.7) over pseudo-LiDAR). Ablation study also demonstrates that our adaptive zooming strategy brings an improvement of over 10\% on AP\textsubscript{3d} (IoU=0.7). In addition, since the official KITTI benchmark lacks fine-grained annotations like pixel-wise part locations, we also present our KFG dataset by augmenting KITTI with detailed instance-wise annotations including pixel-wise part location, pixel-wise disparity, etc.. Both the KFG dataset and our codes will be publicly available at \url{https://github.com/detectRecog/ZoomNet}.

\end{abstract}

\section{Introduction}
Accurate 3D object detection is essential and indispensable for many applications, such as robotics and autonomous driving. Currently, leading 3D object detection methods \cite{yang2019std,shi2019part} for autonomous driving rely heavily on LiDAR (light detection and ranging) data for acquiring accurate depth information. As LiDAR is known to be expensive and has a relatively short perception range (typically around 100m for autonomous driving), stereo cameras which are orders of magnitude cheaper than LiDAR serve as a promising alternative and receive wide attention in both academia and industry. Stereo cameras, which work in a manner similar to human binocular vision, can operate at a higher scanning rate and provide abundant textures cues in high-resolution stereo images for distant objects.

\begin{figure}[!t]
\centering
\includegraphics[width=1.0\linewidth]{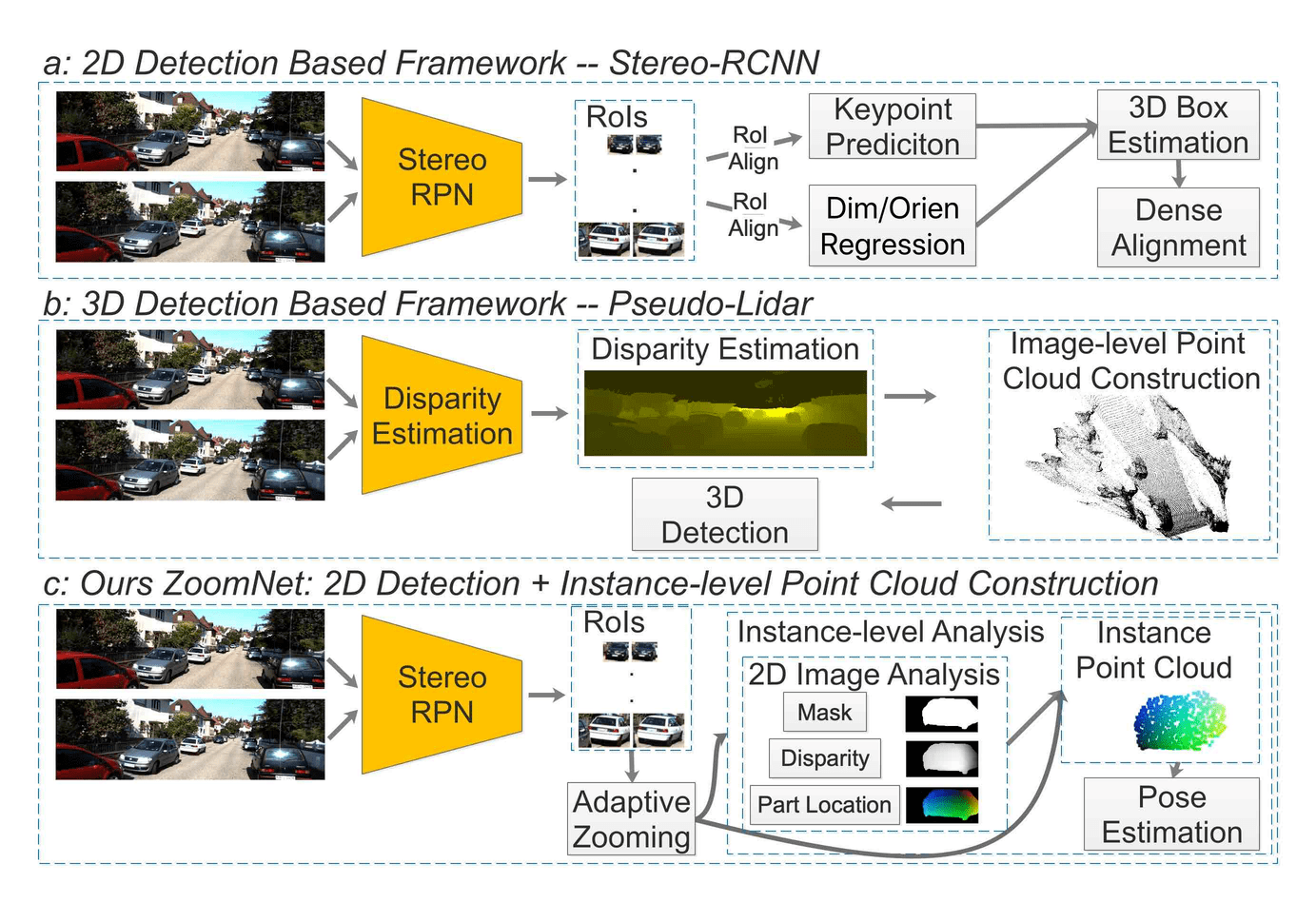}
\caption{Comparisons between ZoomNet and existing frameworks. Framework (a) performs 2D detection then infer 3D pose by 3D-2D geometry constraints. Framework (b) estimates the disparity map to construct pseudo LiDAR, then performs 3D detection models on pseudo LiDAR. By contrast, our proposed ZoomNet, as shown in (c), performs instance-level disparity estimation and analysis, yielding superior performance.}
\label{framework}
\end{figure}

Recent stereo imagery-based 3D detection frameworks put great efforts to explore how stereo 3D information can be effectively used and adapt existing 2d/3d detection models for plugging into their frameworks. As shown in Fig. \ref{framework}a, recent Stereo-RCNN \cite{licvpr2019} first adapts the popular RPN (region proposal networks) for simultaneous 2d detection and box association. The concatenated left-right RoI features are used to regress 2D key-points and 3D dimensions. Then, the 3D pose is estimated by solving 3D-2D projections that are constrained by the 3D box corners with corresponding 2D left-right boxes and key-points. However, such geometry constrain-based method usually rely on high quality bounding box and precise key-point localization which are difficult to guarantee due to the commonly occurred occlusions.
More recently, as shown in Fig. \ref{framework}b, the state-of-the-art framework Pseudo-LiDAR \cite{wang2019pseudo} achieved remarkable progress by converting the representation of stereo depth from disparity maps to pseudo-LiDAR point clouds (PCs) and leveraging the 3D detection models.
The image-level disparity map is estimated in a holistic manner in Pseudo-LiDAR. However, since the visible pixels are limited by image resolution and occlusions, it is usually hard to exam detailed relative positions between stereo image pair and estimate accurate disparity for distant objects. As a result, Pseudo-LiDAR appears weak performance on distant objects. A comparison between Pseudo-LiDAR and our method is shown in Fig. \ref{az_result}.

Naturally, human's vision system can accurately locate objects in 3D spaces via attentively watching and analysing each object instance. It can also hallucinate 3D poses for object with severe occlusion based on the prior knowledge of objects' 3D shapes. Such instance-level analysis and shape priors are far from well exploited in existing stereo imagery-based frameworks.

\begin{figure}[!t]
\centering
\includegraphics[width=0.8\linewidth]{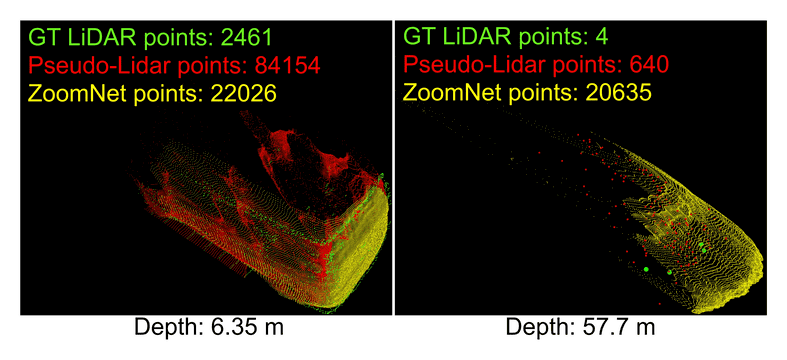}
\caption{By adaptive zooming, the density of our estimated point cloud is uniform no matter how distant the car is. By comparison, the density of pseudo LiDAR and the GT LiDAR drops dramatically. Moreover, our generated points (in blue) are basically distributed smoothly on the visible surface of the car. Best viewed in color.}
\label{az_explain}
\end{figure}

Considering above issues, in this paper, we present a novel framework, as shown in Fig. \ref{framework}c, named Part-aware Adaptive Zooming neural network (ZoomNet). ZoomNet inherits while extends Pseudo-LiDAR by estimating disparity map and constructing PC for each object instance rather than the full image. To realize high quality instance-level disparity estimation, 
we propose a novel strategy named adaptive zooming which fully exploits the convenient zoom in/out on stereo images. 
In adaptive zooming, all RoIs in 2D detection results are adaptively zoomed to the same resolution and camera intrinsic parameters are changed accordingly for projecting pixels on zoomed images to the instance 3D PC. Through adaptive zooming, faraway cars are analyzed at larger resolutions to obtain better disparity estimation and have more compact PCs (see Fig. \ref{az_explain}). Moreover, we introduce the pixel-wise part location which indicates the relative 3D position in the corresponding object for better 3D detection performance on occlusion cases (see Table \ref{occluded_comparison}). Further, several previous 2D detection based methods \cite{licvpr2019,mousavian20173d} utilize 2D classification probability as the 3D detection confidence. We argue that the 3D detection quality, quantified as the 3D IoU between the predicted 3D position and its corresponding ground truth (GT), is not well correlated with the 2D classification probability. Therefore, we put forward the 3D fitting score as a supplementary to the 2D classification probability for better quality estimation. On the popular KITTI 3D detection benchmark, our ZoomNet surpasses all state-of-the-art methods by large margins. Specifically, ZoomNet first reaches a high average precision (AP) of 70\% in the Hard set on AP\textsubscript{3d} (IoU=0.5).

Since the official KITTI 3D detection benchmark only provides the instance 3D position annotation, we introduce KITTI fine-grained car (KFG) dataset to fill the vacancy of fine-grained annotations and to enable the training of ZoomNet. Due to limited space, we leave the details of KFG in the supplementary material.

We summarize our main contributions as follows:
\begin{itemize}
    \item We put forward ZoomNet, a novel stereo imagery-based 3D detection framework.
    \item We present adaptive zooming as an effective strategy by which all instances are analyzed at the same resolution and constructed instance PCs has a uniform density regardless of the instance distance.
    \item We propose part location to improve the robustness of the network against occlusion and 3D fitting score for better 3D detection quality estimation.
    \item Evaluations on KITTI dataset show that ZoomNet outperforms all existing stereo imagery-based methods by large margins. More importantly, it first reaches a high AP of 70\% in the Hard set.
\end{itemize}

\section{Related Work}
We briefly review two stereo imagery-based 3D object detection frameworks and recently-proposed neural networks related to part locations.

\textbf{2D detection based.} 3DOP \cite{chen20173d} is a pioneering work on stereo imagery based 3D object detection. It encodes much information into an energy function and use it to generate 3D proposals for 3D location parameter regression. In \cite{li2018stereo}, the authors propose a semantic tracking system which converts the complex 3D object detection problem into 2D detection, view-point classification, and straightforward closed-form calculation. Triangulation Learning Network \cite{qin2019triangulation} performs 3D detection by exploiting 3D anchors to construct instance correspondences between stereo images. Recently, \citeauthor{licvpr2019} extends Faster-RCNN to simultaneously detect and associate objects in stereo images and utilizes the key-points and 2D constraints to estimate an initial 3D position. After that, the dense alignment strategy is applied to enhance depth estimation significantly. However, dense alignment relies on regions between two boundary key-points. When the lower part of an instance is occluded, this instance will be discarded. By comparison, our proposed part locations are more robust to occlusion than key-points.

\textbf{3D detection based.} With the rapid development of LiDAR processing techniques which handle LiDAR data directly, \citeauthor{wang2019pseudo} propose the pseudo LiDAR representation obtained by computing dense disparity map and then converting them to pseudo LiDAR representations mimicking LiDAR signal. Recently many monocular image-based methods \cite{ma2019accurate,manhardt2019roi,weng2019monocular} also adopt this representation to achieve high 3D detection performance. After that, the best existing LiDAR-based 3D detection algorithm is adopted in order to detect cars. Converting 2D image pixels to 3D PCs can ease the severe occlusion in 2D images and thus makes detection much easier. Therefore, to date, 3D detection based methods \cite{wang2019pseudo,you2019pseudo} achieve higher performances. However, as the depth is calculated at the image level, the estimated depth of distant cars are error-prone. In this paper, we propose adaptive zooming as an effective strategy to obtain better 3D detection results for distant cars.

\textbf{Part-aware neural networks.} As cars' body is rigid, intro-object part locations are informative representations for estimating the pose of cars. Different from key-points, every visible pixel of a car is a valid part location. Recently, in a LiDAR-based method \cite{shi2019part}, the authors also utilize part annotations to help optimize the 3D detection result. However, part locations are still never explored in image-based methods.

% overall pipeline
\begin{figure}[!t]
\centering
\includegraphics[width=0.8\linewidth]{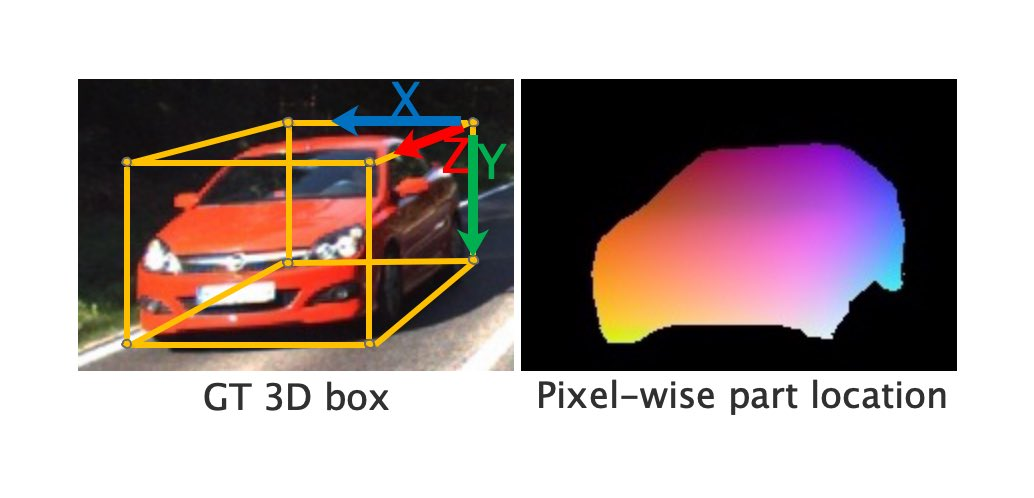}
\caption{Illustration of the formulation of part locations. We regard the pixel-wise part location as RGB channels for visualization.}
\label{part_explain}
\end{figure}

\section{Methodology}

As shown in Fig. \ref{paznet_pipeline}, our proposed ZoomNet contains two major stages: (A) 2D detection and adaptive zooming. (B) instance-wise fine-grained analysis. In stage (A), we perform a joint 2D object detection and bounding box association on the input stereo images. For each instance detected, a 2D classification probability and sizes of 3D bounding box (or dimension $M$) are estimated. In stage (B), ZoomNet performs a fine-grained analysis for all instances, where pixel-wise estimations are predicted, including disparity, part location, foreground/background (f/g) labeling. Then, instance-wise PCs are constructed for pose estimation utilizing these pixel-wise estimations. In pose estimation, the rotation $R$, translation $T$, and a 3D fitting score measuring the goodness of a 3D bounding box are predicted. Hence, the final output of our ZoomNet for a instance is $(M, R, T, S)$, where the 3D detection confidence $S$ is the product of the 2D classification probability and the 3D fitting score.

\subsection{2D Detection and Adaptive Zooming}
In this stage, given a paired left (denoted as $\mathcal{L}$) and right (denoted as $\mathcal{R}$) images, the object detector predicts a list of stereo instances and their corresponding dimensions $M$ and 2D classification probability $P$. Then, by adaptive zooming, faraway cars are analyzed on larger resolutions for better depth estimation.

\subsubsection{2D Detection.}
Different from common 2D detection tasks, object detection on stereo images needs to associate region proposals in stereo images. Recently, Stereo-RCNN \cite{licvpr2019} extends Faster-RCNN for stereo inputs to simultaneously detect and associate instance in stereo images. Then, the stereo RoI features are extracted and are concatenated to classify object categories and to regress dimensions, orientations, as well as key-points. We adopt Stereo-RCNN as the backbone for 2D detection and abandon the branches for key-points and orientations. An instance RoI containing a pair of stereo boxes can be represented by $(x, \bar{x}, y, w, h)$ where $(x,y,w,h)$ denotes the left RoI $L$ ($(x,y), w,h$ is the left-up corner coordinate, the width, and the height respectively) and $(\bar{x},y,w,h)$ denotes the corresponding right RoI $R$.

\begin{figure*}[!t]
\centering
\includegraphics[width=2.0\columnwidth]{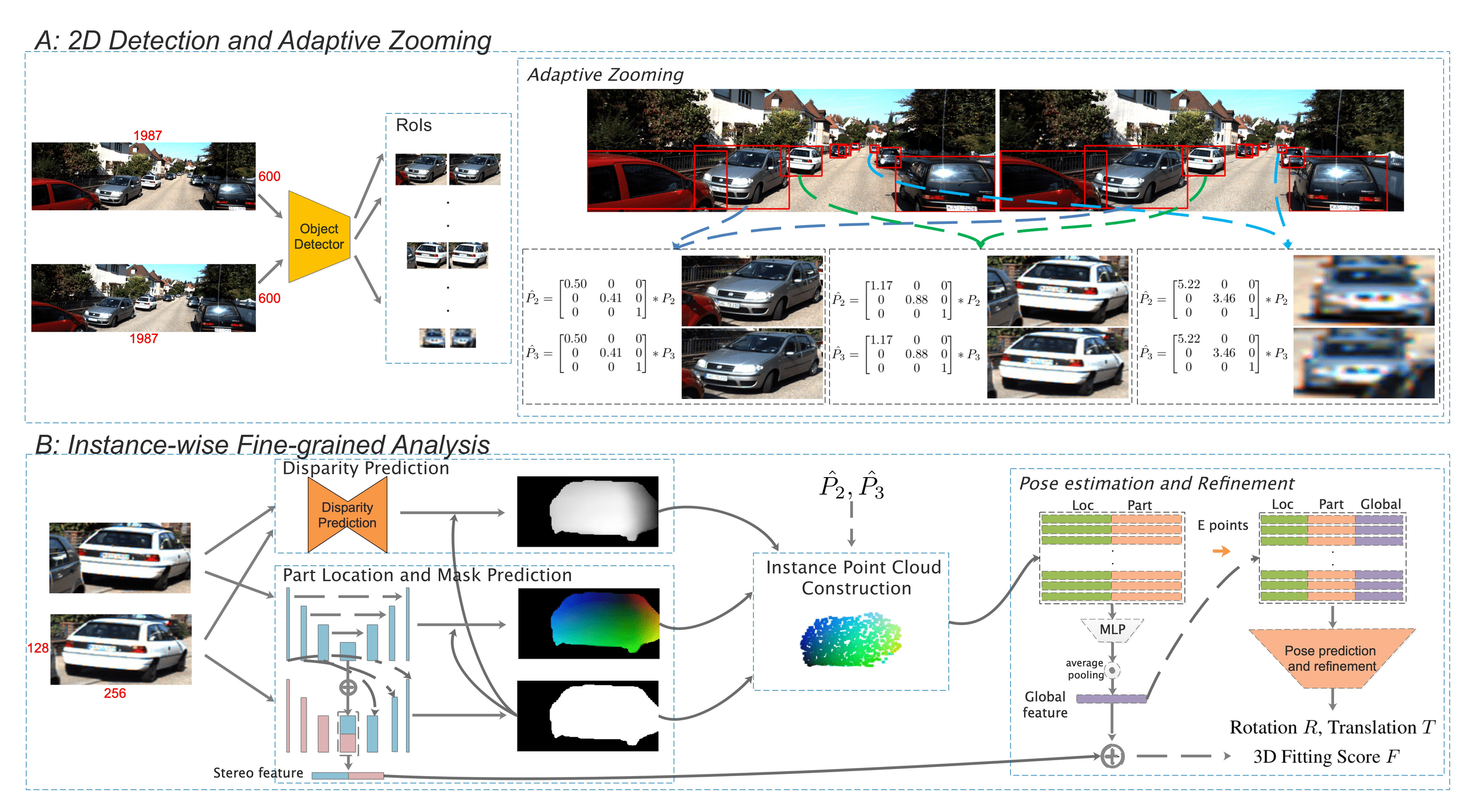}
\caption{Overview of our proposed ZoomNet. In stage (A), ZoomNet first detects and associates instances from stereo images. Then, it scales all instances to the same dimension and adjusts the camera intrinsics. In stage (B), the disparity map, part location map as well as the segmentation map are jointly estimated. Then, these 2D feature maps are converted to 3D instance-wise point clouds for subsequent pose estimation and 3D fitting score regression.
Best viewed in color.}
\label{paznet_pipeline}
\end{figure*}

\subsubsection{Adaptive Zooming.} 
To obtain better depth estimation for distant cars, we introduce an effective strategy named adaptive zooming (AZ). When we sample pixels from film coordinates (projected image), pixels may be rectangular instead of square \cite{forsyth2002computer}. Therefore, the size of the resulting pixel array can vary with two zooming parameters: the horizontal zooming parameter $k$ and the vertical zooming parameter $m$. Following the official KITTI notations \cite{geiger2013vision}, we denote the left camera and right camera as $P_{2}$ and $P_{3}$ respectively. By choosing different zooming parameters to re-sample pixels, the intrinsics of $P_{2}$ change as follows:
{\scriptsize\begin{equation}
\hat{P_2} =\!\!
    \begin{bmatrix}
    k & 0 & 0 \\
    0 & m & 0 \\
    0 & 0 & 1 \\
  \end{bmatrix} 
  \! * P_2 = \!
   \begin{bmatrix}
    k f_{u}^{(2)} & 0 & k c_{u}^{(2)} & -k f_{u}^{(2)} b_{x}^{(2)} \\
    0 & m f_{v}^{(2)} & m c_{v}^{(2)} & 0 \\
    0 & 0 & 1 & 0 \\
  \end{bmatrix} 
\end{equation}\par}

where the superscript $(2)$ indicates it belongs to $P_{2}$. $f_u, f_v$ represents the focal length and $c_u, c_v$ denotes the principal point offset. Besides, $b_x$ denotes the baseline with respect to the reference camera. The intrinsic parameters of the right camera $P_{3}$ change to $\hat{P_3}$ in the same way. $k$ and $m$ are zooming parameters adjusting the focal length of $P_2$ and $P_{3}$. Besides, the zooming parameters also adjust the size of the left RoI $L$ and right RoI $R$ such that the size of resulting images equals to the expected size in the second stage, that is: $ k = \mathcal{W}/w, m = \mathcal{H}/h$ , where $(\mathcal{W},\mathcal{H})$ is the pre-defined input image size in stage (B).
%Besides, the left RoI $L$ and the right RoI $R$ are zoomed to the same resolution for analysis in stage (B).

Different from the depth estimation in Pseudo-LiDAR based methods \cite{wang2019pseudo,you2019pseudo}, AZ can reduce the error of depth estimation by $k$ for distant cars by analyzing the disparity on a larger image. Suppose the image is zoomed horizontally by $k$, for a given error in disparity $\delta D$, AZ influences the error in depth $\delta Z$ as follows:
\begin{equation}
    %Z \propto \frac{1}{D} &\Rightarrow \delta Z \propto \frac{1}{D^2} \delta D \Rightarrow \delta Z \propto Z^2 \delta D, &\text{in P-L} \\
    Z \propto \frac{k}{D}  \Rightarrow \delta Z \propto \frac{k}{D^2} \delta D \Rightarrow \delta Z \propto \frac{1}{k} Z^2 \delta D
    \label{eq1}
\end{equation}
The effectiveness of AZ for depth estimation is also demonstrated by comparative experiments on cars at different distances (see Fig. \ref{az_result}).

\subsection{Instance-wise Fine-grained Analysis}
In this stage, ZoomNet performs fine-grained analysis for all instances. We denote the zoomed left RoI and zoomed right RoI as $\hat{L}$ and $\hat{R}$ respectively. For a left input $\hat{L}$, three pixel-wise predictions are generated, i.e., a disparity map, a part location map, and a f/g labeling. Then, a 3D pose is estimated based on these three predictions. Finally, the goodness of a 3D bounding box estimated is scored by considering both the 2D classification probability and our proposed 3D fitting score. Thanks to the versatile and flexible deep neural networks, we design an end-to-end trainable networks to accomplish all process, which consists of three sub-networks.

\subsubsection{Disparity Prediction}
Following Pseudo-LiDAR, we adopt PSMnet \cite{chang2018pyramid} to perform stereo matching between $\hat{L}$ and $ \hat{R}$ with a smaller maximum disparity value because $\hat{L}$ and $\hat{R}$ are roughly aligned.
Different from Pseudo-LiDAR which fixes the network parameters after training, gradients are permitted to proceed to enable end-to-end learning in out ZoomNet. We think that end-to-end learning allows the sub-network to be trained with additional supervisions from subsequent 3D rotation and translation regression, enabling the network to predict disparities more accurately.

\subsubsection{Part Location and Mask Prediction}
We formulate the part label of a foreground pixel as three continuous values $(p_x, p_y, p_z) \in [0,1]$ to indicate its relative 3D position in object coordinate. Different from \cite{shi2019part}, our part locations are predicted on 2D pixels rather than 3D points. As shown in Fig. \ref{part_explain}, we visualize the part locations by regarding $(p_x, p_y, p_z)$ as three RGB channels.

We design an encoder-decoder architecture for estimating part location and f/g segmentation mask on $L$ (see Fig. \ref{paznet_pipeline}). The encoder is shared, but we use different decoders for part locations and b/f segmentation mask. As predicting part locations on $L$ is straightforward, it is ambiguous to decide the interested instance purely on $L$ when two cars overlap heavily. Thus, as shown in Fig. \ref{paznet_pipeline}, we aggregate the features of $L$ and the features of $R$ extracted by the same encoder in mask prediction to alleviate this ambiguity.

\subsubsection{Point Cloud Construction}
Given the segmentation mask, the part location map, and the left disparity map, we construct PCs from foreground pixels. Converting a pixel in the image plane to a 3D point in the camera coordinate system needs the pixel coordinate on the 2D image as well as the corresponding camera intrinsic parameters. A common practice is reshaping these 2D feature maps of the height $kw$ and the width $mh$ back to the original width $w$ and height $h$, then use the original camera intrinsic parameters $P_2$ and $P_3$ to project pixels in the original input image $\mathcal{L}$ to 3D points. With the help of AZ, we do not need to down-sample or up-sample 2D feature maps. Instead, we perform this projection on the hallucinated zoomed left image $\hat{\mathcal{L}}$ using the corresponding pseudo camera intrinsic parameters $\hat{P_2}$ and $\hat{P_3}$. $\hat{\mathcal{L}}$ is obtained by scaling $\mathcal{L}$ by $k$ in the horizontal direction and by $m$ in the vertical direction. In this way, the starting point of $L$ changes from $(x,y)$ to $(kx, my)$ and the disparity offset between $L$ and $R$ changes to $\hat{O} = k (x - \bar{x})$. However, the baseline distance $\hat{b_{l}}$ between $\hat{P_2}$ and $\hat{P_3}$ stays unchanged:
\begin{equation}
\begin{split}
    \hat{b_{l}} = (-k f_{u}^{(2)} b_{x}^{(2)} + k f_{u}^{(3)} b_{x}^{(3)}) / (k f_{u}^{(2)}) \\
    = (-f_{u}^{(2)} b_{x}^{(2)} + f_{u}^{(3)} b_{x}^{(3)}) / (f_{u}^{(2)})
 \end{split}
\end{equation}

For each foreground point $(u,v), u \in [0, \mathcal{W}), v \in [0,  \mathcal{H})$ on the segmentation map, we denote the predicted disparity of this point as $d_{u,v}$. The position of a 3D point in camera coordinate, $(x, y, z)$ can be computed as:
\begin{equation}
    \begin{cases}
  x=  (u + kx -k c_{u}^{(2)}) * z / (k f_{u}^{(2)}) + b_{x}^{(2)}\\
  y=  (v + my -m c_{v}^{(2)}) * z / (m f_{v}^{(2)})\\
  z= (k f_{u}^{(2)} \hat{b_{l}}) / (d_{u,v} + \hat{O})
  \end{cases}
\end{equation}

Also, for each 3D point $(x,y,z)$ in camera coordinate, the part location $(p_x, p_y, p_z)$ in object coordinate is also integrated. Suppose there are $N$ foreground pixels on the segmentation map, then the dimension of the constructed PC is $N*6$, where $N*3$ is the 3D location and another $N*3$ is the corresponding estimated part location.

Thanks to adaptive zooming, the constructed instance PC usually has a uniform density (see Fig. \ref{az_explain}) no matter the distance of the car is near or far. In ablation study (see Table \ref{ablation}), we will show that the high density PC is beneficial for high quality 3D detection.

\subsubsection{Pose Estimation and Refinement}
After the instance-wise PC is constructed, we sample a pre-defined number $E$ of points for 3D feature extraction. As points generated by ZoomNet are basically evenly distributed on the instance surface as discussed before, we simply sample $E$ points randomly. The common practice is to estimate the rotation $R$ and the translation $T$ by a direct regression or a combination of classification and regression \cite{shin2018roarnet}. However, inspired by a recent work named DenseFusion \cite{wang2019densefusion} on RGB-D pose estimation, we think that a CAD model-based distance loss can better measure the quality of pose estimation. We hence modify DenseFusion for our pose estimation and tailor it by removing its CNN module and exploiting predicted part locations as the color embeddings. 

As aforementioned, the 2D classification probability $P$ is not well correlated with the 3D detection quality. Therefore, we propose the 3D fitting score to better measure the quality of 3D detection. However, predicting the quality purely on the constructed PC is very ambiguous. Therefore, we introduce the stereo imagery features because the stereo images features provide many clues related to the 3D detection quality like the distance, the occlusion level, etc. The stereo imagery features and the global PC features are aggregated and are feed to a fully connected layer for 3D fitting score regression. The training objective $\hat{F}$ is associated with the mean distance $\hat{D}$ between the constructed instance PC and the GT instance PC constructed by the GT disparity. For each foreground point (take $n$ for example), let $[X_n,Y_n,Z_n]$ be the generated point coordinate and $[\hat{X_n},\hat{Y_n},\hat{Z_n}]$ be the GT point coordinate. The training target is formulated as:
\begin{equation}
\begin{split}
     \hat{D} &= \textnormal{Clip}(\frac{1}{E} \sum_{n} |Z_n-\hat{Z_n}|) \in [0.0, 1.0]\\
     \hat{F} &= 1.0 - e^{- \frac{\hat{D}}{\theta}}
\end{split}
\label{score}
\end{equation}

We clip $\hat{D}$ to the range $[0.0,1.0]$ as the 3D detection result often fails when the depth error exceeds 1 meter. Since it is hard to estimate depth correctly, we specifically design the loss to enforce the network to focus on depth prediction by using Eq. (\ref{score}).

\subsection{Training}
We leave the training details including stereo-RCNN and pose estimation in the supplementary material. In stage (B), the input image size $(\mathcal{W},\mathcal{H})$ is set to $(256, 128)$. $E$ is set to $500$ in both training and testing phases because empirically we find sampling more points does not enhance the pose estimation performance.
The multi-task loss in stage (B) is formulated as follows:
\begin{equation}
    L = L_{m} + L_{d} + L_{pa} + L_{po} + L_{r} + L_{s}
\end{equation}
where $L_{m}$ is the binary cross-entropy loss between the estimated pixel-wise segmentation score and the GT b/f segmentation mask. For each GT foreground pixel, $L_{d}$ is the smooth L1 distance between the estimated disparity and the GT disparity, and $L_{pa}$ is the smooth L1 distance between the estimated part location and the GT part location on the GT foreground area. $L_{po}$ and $L_{r}$ are CAD model-based pose estimation loss and refinement loss (refer to \cite{wang2019densefusion} for more details), respectively. $L_{s}$ is the smooth L1 distance between the predicted 3D fitting score $F$ and $\hat{F}$. Empirically we find the 3D fitting score cooperates with the 2D classification probability better when $\theta$ is set to $8$. Moreover, to stabilize the joint training, a lower weight $w=0.1$ is assigned to the loss of part location and mask. The learning rate in pose estimation and refinement is set to $0.1$ of the global learning rate. 

During training in stage (B), we perform data augmentation by randomly but ``moderately" shifting the horizontal starting point of the left/right crops. By default, the left/right crops are cropped at the left/right bounding box of the 2D detection result. When the horizontal starting points $\bar{x},x$ of the left/right bounding box change, the pixel-wise disparity between the Left/Right crop changes accordingly.``Moderately" shifting means that the resulting disparity map should satisfy two requirements: (1) the minimal disparity of foreground pixels should be greater than zero; (2) the maximum disparity of foreground pixels should be smaller than the default maximum disparity for stereo matching.

\begin{table*}[!t]
\resizebox{2.0\columnwidth}{!}{%
\begin{tabular}{@{}|l|c|cccccc|cccccc|@{}}
\hline
 &  & \multicolumn{3}{c|}{AP\textsubscript{bv} (IoU=0.5)} & \multicolumn{3}{c|}{AP\textsubscript{bv} (IoU=0.7)} & \multicolumn{3}{c|}{AP\textsubscript{3d} (IoU=0.5)} & \multicolumn{3}{c|}{AP\textsubscript{3d} (IoU=0.7)} \\ \cmidrule(l){3-14} 
Method & Detector & Easy & Mode & \multicolumn{1}{c|}{Hard} & Easy & Mode & Hard & Easy & Mode & \multicolumn{1}{c|}{Hard} & Easy & Mode & Hard \\ \midrule
3DOP & 2D & 55.04 & 41.25 & \multicolumn{1}{c|}{34.55} & 12.63 & 9.49 & 7.59 & 46.04 & 34.63 & \multicolumn{1}{c|}{30.09} & 6.55 & 5.07 & 4.10 \\ \midrule
DirectShape & 2D & 54.35 & 38.15 & \multicolumn{1}{c|}{32.03} & 15.55 & 10.77 & 9.22 & 47.62 & 31.01 & \multicolumn{1}{c|}{28.26} & 8.81 & 5.56 & 5.37 \\ \midrule
MLF & 2D & - & 53.56 & \multicolumn{1}{c|}{-} & - & 19.54 & - & - & 47.42 & \multicolumn{1}{c|}{-} & - & 9.80 & - \\ \midrule
TLN & 2D & 62.46 & 45.99 & \multicolumn{1}{c|}{41.92} & 29.22 & 21.88 & 18.83 & 59.51 & 43.71 & \multicolumn{1}{c|}{37.99} & 18.15 & 14.26 & 13.72 \\ \midrule
Stereo-RCNN & 2D & 87.13 & 74.11 & \multicolumn{1}{c|}{58.93} & 68.50 & 48.30 & 41.47 & 85.84 & 66.28 & \multicolumn{1}{c|}{57.24} & 54.11 & 36.69 & 31.07 \\ \midrule
Pseudo-LiDAR & 3D & 89.00 & 77.50 & \multicolumn{1}{c|}{68.70} & 74.90 & 56.80 & 49.00 & 88.50 & 76.40 & \multicolumn{1}{c|}{61.20} & 61.90 & 45.30 & 39.00 \\ \midrule
ZoomNet & 2D & \textbf{90.62} & \textbf{88.40} & \multicolumn{1}{c|}{\textbf{71.44}} & \textbf{78.68} & \textbf{66.19} & \textbf{57.60} & \textbf{90.44} & \textbf{79.82} & \multicolumn{1}{c|}{\textbf{70.47}} & \textbf{62.96} & \textbf{50.47} & \textbf{43.63} \\ \bottomrule
\end{tabular}%
}
\caption{Average precision of bird's-eye-view (AP\textsubscript{bv}) and 3D boxes (AP\textsubscript{3d}) comparison, evaluated on the KITTI validation set.}
\label{main_result}
\end{table*}

\section{Experiments}
We evaluate our method on the challenging KITTI object detection benchmark \cite{geiger2013vision}. The main results are shown in Table \ref{main_result}, where we compare ZoomNet with previous state-of-the-arts. To demonstrate the effectiveness of adaptive zooming, we carry out comparative experiments on cars at different distances. Moreover, We test ZoomNet on different occlusion levels to examine its resistance against occlusion. Lastly, we perform ablation study to provide a deeper understanding of ZoomNet. More details of training ZoomNet can be found in the supplementary material.

\subsection{Metric}
We focus on 3D object detection (AP\textsubscript{3d}) and bird's-eye-view (BEV) object detection (AP\textsubscript{bv}) and report the results on the validation set. AP\textsubscript{3d} and AP\textsubscript{bv} are the average precision measured in 3D and BEV, respectively. Following previous works \cite{wang2019pseudo,licvpr2019,chen20173d}, we focus on the `car' category and split the 7481 training images into training set and validation set with roughly the same amount. It is worth noting that all cars are divided into three difficulties: Easy, Mode (moderate) and Hard according to different bounding box heights and different occlusion/truncation levels. Moreover, there is a category named `DontCare' which denotes regions where objects have not been labeled, and detection results on these regions will be ignored. Therefore, when we compare methods on a specific condition, by default, we set the category of all cars that do not meet this condition as `DontCare'.

\subsection{3D and BEV object detection}
We compare recent works on stereo imagery-based 3D detection: 3DOP \cite{chen20173d}, MLF \cite{xu2018multi}, DirectShape \cite{wang2019directshape}, TLN \cite{qin2019triangulation}, Stereo R-CNN \cite{licvpr2019}, and Pseudo-LiDAR \cite{wang2019pseudo}. Except for Pseudo-LiDAR, all methods are based on 2D detection. Pseudo-LiDAR provides multiple 3D object detectors for chosen. We select the one with the highest performance. The recently improved version of Pseudo-LiDAR \cite{you2019pseudo} is not included because its code is not publicly available yet.

\begin{figure*}[!t]
  \centering
  \includegraphics[width=0.8\columnwidth]{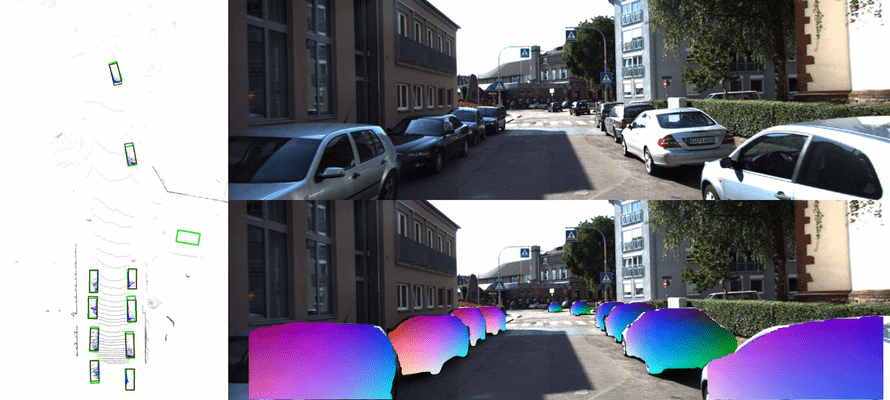}
\includegraphics[width=0.8\columnwidth]{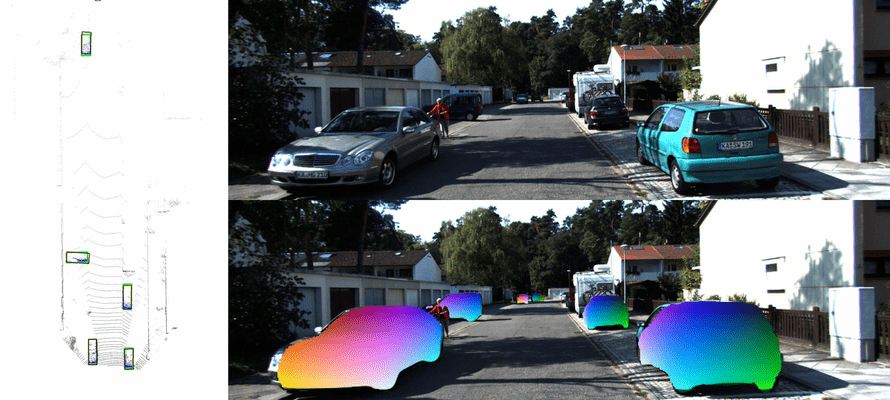}
\includegraphics[width=0.8\columnwidth]{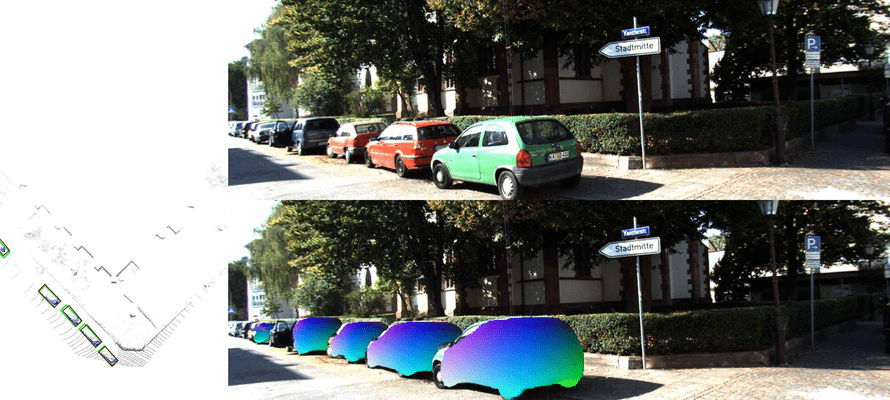}
\includegraphics[width=0.8\columnwidth]{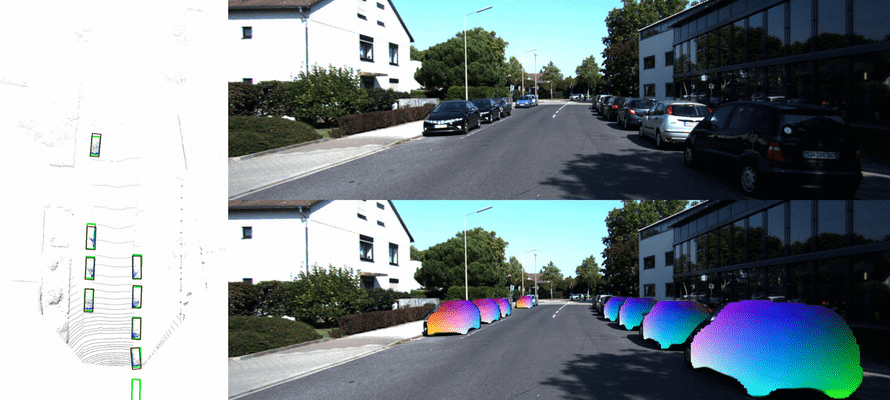}
\caption{Quantitative results. For each example, the image on the left is the BEV map, the two stacked images on the right are the original left image and the visualization of predicted part locations on predicted masks. On the BEV image, the red box represents the detection box and the green one indicates the GT box. In addition, the GT LiDAR points are shown in gray while points generated by ZoomNet are shown in blue.}
\label{sample}
\end{figure*}

We summarize the main results in Table \ref{main_result}, where our method outperforms all state-of-the-art methods, especially on the AP\textsubscript{bv} at the IoU thresh of 0.7. Quantitative results are shown in Fig. \ref{sample}. While adopting the same detector, our method outperforms Stereo R-CNN by large margins on both AP\textsubscript{bv} and AP\textsubscript{3d}. On the most challenging metric AP\textsubscript{3d} (IoU=0.7), we steadily surpass Stereo R-CNN by over 14\% in the Mode and Hard set. Though we only surpass Pseudo-LiDAR by a small margin on AP\textsubscript{3d} (IoU=0.7) in the Easy set, we report a large improvement of over 9\% on AP\textsubscript{3d} (IoU=0.5) in the most difficult Hard set. More importantly, in the Hard set of AP\textsubscript{3d} (IoU=0.5), our method achieves a high AP of 70.47\%. While LiDAR-based detectors do not have the problem of accurate depth estimation, this is the first time that stereo imagery based method achieves over 70\% AP in the challenging Hard set on AP\textsubscript{3d} (IoU=0.5). 

How can ZoomNet achieve a nearly 70\% AP in the challenging Hard set? Since cars in the Hard set are mostly distant and are difficult to see, the reasons may be two-fold: (i) Thanks to adaptive zooming, ZoomNet can detect distant cars more accurately by introducing a distance-invariant strategy. (ii) ZoomNet has strong robustness to occlusion. In the following subsections, we investigate the 3D detection performance at different distances and compare ZoomNet with on different occlusion levels.

\subsection{Comparisons at Different Distances}
\begin{figure}[!t]
\centering
\includegraphics[width=0.95\linewidth]{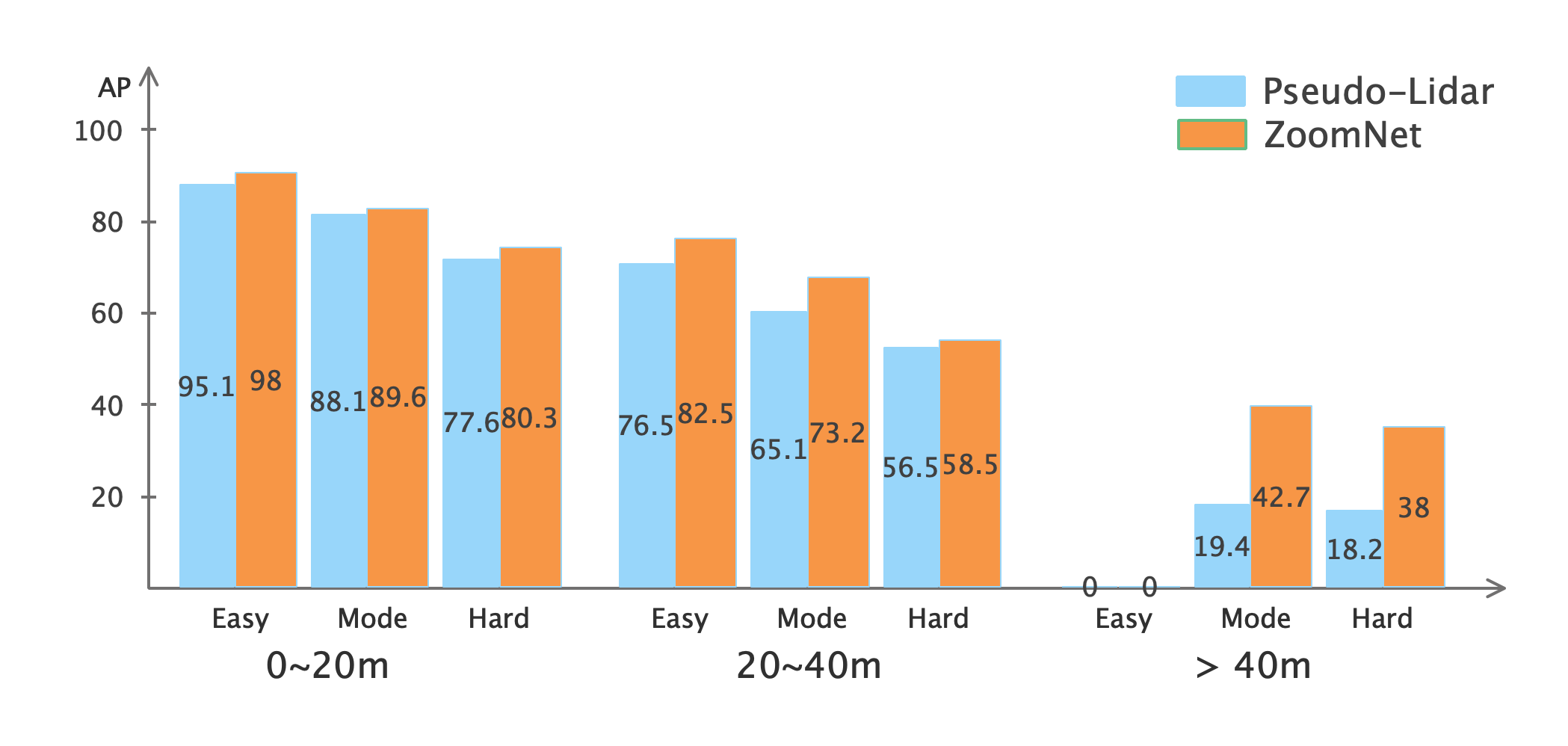}
\caption{Comparisons between ZoomNet and Pseudo-LiDAR at different distances on AP\textsubscript{3d} (IoU=0.5).}
\label{az_result}
\end{figure}

\begin{table}[!t]
\centering
\resizebox{0.8\columnwidth}{!}{
\begin{tabular}{|c|c|c|c|}
\hline
      & Stereo R-CNN & ZoomNet &  ZoomNet (no part)       \\ \cline{2-4}
Level & Hard        & Hard   & Hard       \\ \hline
0     & 43.12       & 51.73  & 46.69             \\ \hline
1     & 17.90       & 34.99  & 28.16               \\ \hline
2     & 6.21       & 9.76  & 5.56           \\ \hline
\end{tabular}
}
\caption{Comparisons on different levels of occlusion on AP\textsubscript{3d} (IoU=0.7).}
\label{occluded_comparison}
\end{table}

Compared with estimating full image disparity, as explained in Eq. (\ref{eq1}), AZ can reduce the depth estimation error for distant cars. To test the effect of AZ on 3D detection performance at different distances, we divide all cars in the validation set into three categories by depth: (i) $0\sim20$m, (ii) $20\sim40$m, and (iii) $>40$m. The result is shown in Fig. \ref{az_result}. For near cars ($0\sim20$m), the gap between Pseudo-LiDAR and our ZoomNet is small. Their high performance at the near distance also demonstrates that stereo imagery-based methods are promising for low-speed robots because low-speed robots, like we humans, do not need to care about objects that are too far away. 

With the depth increasing to $40$m and more, the performance gap is growing larger and larger. Especially in the Hard set, for cars of more than $40$ meters away, ZoomNet achieves 42.7\% AP in the Mode set, about 120\% higher than Pseudo-LiDAR. This trend validates that ZoomNet has an obviously better 3D detection performance for distant cars.

\subsection{Comparisons on Different Occlusion Levels}
To validate that ZoomNet is more robust to occlusion than key-points based methods, we compare ZoomNet with Stereo R-CNN on different occlusion levels. The result on AP\textsubscript{3d} (IoU=0.7) is shown in Table \ref{occluded_comparison}. Occlusion Level $0, 1, 2$ represent `Fully visible', `Partly occluded', `Difficult to see', respectively. Here we only care about the comparison in the Hard set because cars belonging to occlusion Level $2$ only exist in the Hard set. 
On all occlusion levels, our ZoomNet steadily surpasses Stereo R-CNN by large margins. Especially on the occlusion level $1$, the performance of ZoomNet is 95\% higher than Stereo R-CNN. The performance increment demonstrates that ZoomNet performs better than key-points based methods in occlusion cases. However, when the part location is removed in ZoomNet, the performance on the occlusion level $2$ drops a lot and becomes slightly worse than Stereo R-CNN. The tremendous decrease in AP indicates that part locations are crucial for detecting occluded cars. The importance of part locations is also demonstrated in the following ablation study.

\subsection{Testset Results}

\begin{table}[!t]
\centering
\resizebox{\columnwidth}{!}{
\begin{tabular}{|c|ccc|ccc|}
\hline
             & \multicolumn{3}{c|}{AP\textsubscript{bv} (IoU=0.7)} & \multicolumn{3}{c|}{AP\textsubscript{3d} (IoU=0.7)} \\ \cline{2-7} 
Method       & Easy     & Mode     & Hard     & Easy     & Mode     & Hard     \\ \hline
Stereo-RCNN  & 61.92    & 41.31    & 33.42    & 47.58    & 30.23    & 23.72    \\ \hline
Pseudo-LiDAR & 67.30    & 45.00    & 38.40    & 54.53    & 34.05    & 28.25    \\ \hline
ZoomNet      & \textbf{72.94}    & \textbf{54.91}    & \textbf{44.14}    & \textbf{55.98}    & \textbf{38.64}    & \textbf{30.97}    \\ \hline
\end{tabular}}
\caption{3D detection and localization AP on the KITTI test set.}
\label{testset}
\end{table}

To demonstrate the effectiveness of the proposed model, we also report the evaluation results on the official KITTI testing set in Table \ref{testset}. The detailed performance can be found online. \footnote{\url{http://www.cvlibs.net/datasets/kitti/eval_object_detail.php?&result=b858fba7c42945bb554a56f43c3db682cd475c9e}} It is worth noting that our ZoomNet has surpassed state-of-the-art methods on AP\textsubscript{bv} (IoU=0.7) by 9.9\% in the Mode set.

\subsection{Ablation Study}

\begin{table}[!t]
\centering
\resizebox{\columnwidth}{!}{
\begin{tabular}{|c|c|c|ccc|ccc|}
\hline
     &      &    & \multicolumn{3}{c|}{AP\textsubscript{3d} (IoU=0.5)} & \multicolumn{3}{c|}{AP\textsubscript{3d} (IoU=0.7)} \\ \cline{4-9} 
Part & 3DFS & AZ & Easy     & Mode     & Hard     & Easy     & Mode     & Hard     \\ \hline
-    & -    & -         &85.49      &74.61  & 65.41  &45.80 & 38.33 & 33.01 \\ \hline
-    & $\surd$    & $\surd$  & 88.54    & 77.99    & 68.20    & 54.22    & 40.87    & 34.51    \\ \hline
$\surd$    & -    & $\surd$  & 88.55    & 77.64     & 68.81    & 57.51 &   46.67   &40.88    \\ \hline
$\surd$    & $\surd$    & -  & 87.00    & 76.51    & 67.36    & 52.07    & 40.39    & 34.20    \\ \hline
$\surd$    & $\surd$    & $\surd$  & 90.44  &79.82  &70.47    & 62.96   &50.47  &43.63    \\ \hline
\end{tabular}
}
\caption{Ablation study on ZoomNet. 3DFS denotes 3D fitting score.
}
\label{ablation}
\end{table}

In this ablation study, we validate the contributions of part locations, 3D fitting score (3DFS), and AZ Results are shown in Table \ref{ablation}. When not using part locations, we adjust the input channels of pose estimation. When not using 3DFS, we regard the 2D classification probability as the 3D detection confidence. When not using AZ, we reshape 2D feature maps back to the original size and project pixels in $\mathcal{L}$ rather than on the zoomed image $\hat{\mathcal{L}}$ to construct PCs. 

On AP\textsubscript{3d} (IoU=0.5), the absence of part location or 3DFS leads to a small performance drop of 2\% in the Mode set. However, at the higher 3D IoU threshold 0.7, non-trivial drops of 10\% and 4\% are observed when part location or 3DFS is removed. The small gap on AP\textsubscript{3d} (IoU=0.5) and the relatively large gap on AP\textsubscript{3d} (IoU=0.7) demonstrate that part location and 3DFS are crucial for high-quality 3D detection. 

When AZ is removed, the PC density decreases dramatically for distant cars. For cars which are 55 meters away, the number of points drops from 20k to less than 1k. The performance also drops dramatically. In practice, we observe that the end-to-end training in stage (B) also becomes unstable.

Combining these three ingredients together, large improvements on AP\textsubscript{3d} (IoU=0.7) are observed, and our integrated ZoomNet pipeline achieves the best 3D detection results among current stereo-imagery based methods.

\section{Conclusion}
In this paper, we propose a novel framework for stereo imagery-based 3D detection. By exploiting adaptive zooming to process cars at various distances at a unified resolution and utilizing part locations to solve the ambiguity under severe occlusion, our method ZoomNet surpasses all existing image-based methods by large margins on the challenging KITTI 3D and BEV object detection tasks. More importantly, ZoomNet is the first image-based method to achieve a high AP of 70\% in the Hard set. The high 3D detection performance demonstrates that ZoomNet can serve as a competitive and reliable backup for autonomous driving and robot navigation.

\section{Acknowledgment}
This work was supported by the National Natural Science Foundation of China (No. 61602400) and the Anhui Initiative in Quantum Information Technologies (No. AHY150300).

{\small
\bibliographystyle{aaai}
\bibliography{ref}

\begin{thebibliography}{}

\bibitem[\protect\citeauthoryear{Chang and Chen}{2018}]{chang2018pyramid}
Chang, J.-R., and Chen, Y.-S.
\newblock 2018.
\newblock Pyramid stereo matching network.
\newblock In {\em Proceedings of the IEEE Conference on Computer Vision and
  Pattern Recognition},  5410--5418.

\bibitem[\protect\citeauthoryear{Chen \bgroup et al\mbox.\egroup
  }{2017}]{chen20173d}
Chen, X.; Kundu, K.; Zhu, Y.; Ma, H.; Fidler, S.; and Urtasun, R.
\newblock 2017.
\newblock 3d object proposals using stereo imagery for accurate object class
  detection.
\newblock {\em IEEE transactions on pattern analysis and machine intelligence}
  40(5):1259--1272.

\bibitem[\protect\citeauthoryear{Forsyth and Ponce}{2002}]{forsyth2002computer}
Forsyth, D.~A., and Ponce, J.
\newblock 2002.
\newblock {\em Computer vision: a modern approach}.
\newblock Prentice Hall Professional Technical Reference.

\bibitem[\protect\citeauthoryear{Geiger \bgroup et al\mbox.\egroup
  }{2013}]{geiger2013vision}
Geiger, A.; Lenz, P.; Stiller, C.; and Urtasun, R.
\newblock 2013.
\newblock Vision meets robotics: The kitti dataset.
\newblock {\em The International Journal of Robotics Research}
  32(11):1231--1237.

\bibitem[\protect\citeauthoryear{Li, Chen, and Shen}{2019}]{licvpr2019}
Li, P.; Chen, X.; and Shen, S.
\newblock 2019.
\newblock Stereo r-cnn based 3d object detection for autonomous driving.
\newblock In {\em CVPR}.

\bibitem[\protect\citeauthoryear{Li, Qin, and others}{2018}]{li2018stereo}
Li, P.; Qin, T.; et~al.
\newblock 2018.
\newblock Stereo vision-based semantic 3d object and ego-motion tracking for
  autonomous driving.
\newblock In {\em Proceedings of the European Conference on Computer Vision
  (ECCV)},  646--661.

\bibitem[\protect\citeauthoryear{Ma \bgroup et al\mbox.\egroup
  }{2019}]{ma2019accurate}
Ma, X.; Wang, Z.; Li, H.; Ouyang, W.; and Zhang, P.
\newblock 2019.
\newblock Accurate monocular 3d object detection via color-embedded 3d
  reconstruction for autonomous driving.
\newblock {\em arXiv preprint arXiv:1903.11444}.

\bibitem[\protect\citeauthoryear{Manhardt, Kehl, and
  Gaidon}{2019}]{manhardt2019roi}
Manhardt, F.; Kehl, W.; and Gaidon, A.
\newblock 2019.
\newblock Roi-10d: Monocular lifting of 2d detection to 6d pose and metric
  shape.
\newblock In {\em Proceedings of the IEEE Conference on Computer Vision and
  Pattern Recognition},  2069--2078.

\bibitem[\protect\citeauthoryear{Mousavian \bgroup et al\mbox.\egroup
  }{2017}]{mousavian20173d}
Mousavian, A.; Anguelov, D.; Flynn, J.; and Kosecka, J.
\newblock 2017.
\newblock 3d bounding box estimation using deep learning and geometry.
\newblock In {\em Proceedings of the IEEE Conference on Computer Vision and
  Pattern Recognition},  7074--7082.

\bibitem[\protect\citeauthoryear{Qin, Wang, and
  Lu}{2019}]{qin2019triangulation}
Qin, Z.; Wang, J.; and Lu, Y.
\newblock 2019.
\newblock Triangulation learning network: from monocular to stereo 3d object
  detection.
\newblock {\em arXiv preprint arXiv:1906.01193}.

\bibitem[\protect\citeauthoryear{Shi \bgroup et al\mbox.\egroup
  }{2019}]{shi2019part}
Shi, S.; Wang, Z.; Wang, X.; and Li, H.
\newblock 2019.
\newblock Part-a\^{} 2 net: 3d part-aware and aggregation neural network for
  object detection from point cloud.
\newblock {\em arXiv preprint arXiv:1907.03670}.

\bibitem[\protect\citeauthoryear{Shin, Kwon, and
  Tomizuka}{2018}]{shin2018roarnet}
Shin, K.; Kwon, Y.~P.; and Tomizuka, M.
\newblock 2018.
\newblock Roarnet: A robust 3d object detection based on region approximation
  refinement.
\newblock {\em arXiv preprint arXiv:1811.03818}.

\bibitem[\protect\citeauthoryear{Wang \bgroup et al\mbox.\egroup
  }{2019a}]{wang2019densefusion}
Wang, C.; Xu, D.; Zhu, Y.; Mart{\'\i}n-Mart{\'\i}n, R.; Lu, C.; Fei-Fei, L.;
  and Savarese, S.
\newblock 2019a.
\newblock Densefusion: 6d object pose estimation by iterative dense fusion.
\newblock In {\em Proceedings of the IEEE Conference on Computer Vision and
  Pattern Recognition},  3343--3352.

\bibitem[\protect\citeauthoryear{Wang \bgroup et al\mbox.\egroup
  }{2019b}]{wang2019directshape}
Wang, R.; Yang, N.; Stueckler, J.; and Cremers, D.
\newblock 2019b.
\newblock Directshape: Photometric alignment of shape priors for visual vehicle
  pose and shape estimation.
\newblock {\em arXiv preprint arXiv:1904.10097}.

\bibitem[\protect\citeauthoryear{Wang \bgroup et al\mbox.\egroup
  }{2019c}]{wang2019pseudo}
Wang, Y.; Chao, W.-L.; Garg, D.; Hariharan, B.; Campbell, M.; and Weinberger,
  K.~Q.
\newblock 2019c.
\newblock Pseudo-lidar from visual depth estimation: Bridging the gap in 3d
  object detection for autonomous driving.
\newblock In {\em Proceedings of the IEEE Conference on Computer Vision and
  Pattern Recognition},  8445--8453.

\bibitem[\protect\citeauthoryear{Weng and Kitani}{2019}]{weng2019monocular}
Weng, X., and Kitani, K.
\newblock 2019.
\newblock Monocular 3d object detection with pseudo-lidar point cloud.
\newblock {\em arXiv preprint arXiv:1903.09847}.

\bibitem[\protect\citeauthoryear{Xu and Chen}{2018}]{xu2018multi}
Xu, B., and Chen, Z.
\newblock 2018.
\newblock Multi-level fusion based 3d object detection from monocular images.
\newblock In {\em Proceedings of the IEEE Conference on Computer Vision and
  Pattern Recognition},  2345--2353.

\bibitem[\protect\citeauthoryear{Yang \bgroup et al\mbox.\egroup
  }{2019}]{yang2019std}
Yang, Z.; Sun, Y.; Liu, S.; Shen, X.; and Jia, J.
\newblock 2019.
\newblock Std: Sparse-to-dense 3d object detector for point cloud.
\newblock {\em arXiv preprint arXiv:1907.10471}.

\bibitem[\protect\citeauthoryear{You \bgroup et al\mbox.\egroup
  }{2019}]{you2019pseudo}
You, Y.; Wang, Y.; Chao, W.-L.; Garg, D.; Pleiss, G.; Hariharan, B.; Campbell,
  M.; and Weinberger, K.~Q.
\newblock 2019.
\newblock Pseudo-lidar++: Accurate depth for 3d object detection in autonomous
  driving.
\newblock {\em arXiv preprint arXiv:1906.06310}.

\end{thebibliography}
}

\end{document}